\documentclass[letterpaper, 10 pt, conference]{ieeeconf}  

\IEEEoverridecommandlockouts                              

\usepackage{graphicx} 
\usepackage{times} 
\usepackage{amsmath} 
\usepackage{amssymb}  
\usepackage{algorithm}
\usepackage{algpseudocode}
\usepackage{url}
\usepackage[hidelinks]{hyperref}
\usepackage[utf8]{inputenc}

\title{\LARGE \bf
Grasping Student: semi-supervised learning for robotic manipulation
}

\author{Piotr Krzywicki$^{1, 2}$, Krzysztof Ciebiera$^{1}$, Rafał Michaluk$^{1, 2}$, Inga Maziarz$^{1, 3}$, Marek Cygan$^{1, 4}$\\\\
\small{University of Warsaw$^{1}$, IDEAS NCBR$^{2}$, Warsaw University of Technology$^{3}$, Nomagic$^{4}$}
}

\begin{document}

\maketitle
\thispagestyle{empty}
\pagestyle{empty}

\begin{abstract}
Gathering real-world data from the robot quickly becomes
a bottleneck when constructing a robot learning system for grasping.
In this work, we design a semi-supervised grasping system that, on top of a small sample of robot experience, takes advantage of images of products to be picked, which are collected without any interactions with the robot. 

We validate our findings both in the simulation and in the real world. In the regime of a small number of robot training samples, taking advantage of the unlabeled data allows us to achieve performance at the level of 10-fold bigger dataset size used by the baseline.

The code and datasets used in the paper will be released at https://github.com/nomagiclab/grasping-student.
\end{abstract}

\section{INTRODUCTION}
Robotic grasping is an essential area of research in robotics because it enables robots to interact with objects in the physical world, a capability that is necessary for many practical applications such as manufacturing, warehousing, and healthcare. Robotic grasping involves designing algorithms and hardware that enable a robot to perceive and manipulate objects of various sizes, shapes, and materials. The ability to grasp objects reliably and efficiently is a key determinant of a robot's overall performance and effectiveness and therefore is a critical area of focus in robotics research.

\begin{figure}[thpb]
    \begin{center}
        \includegraphics[scale=0.62]{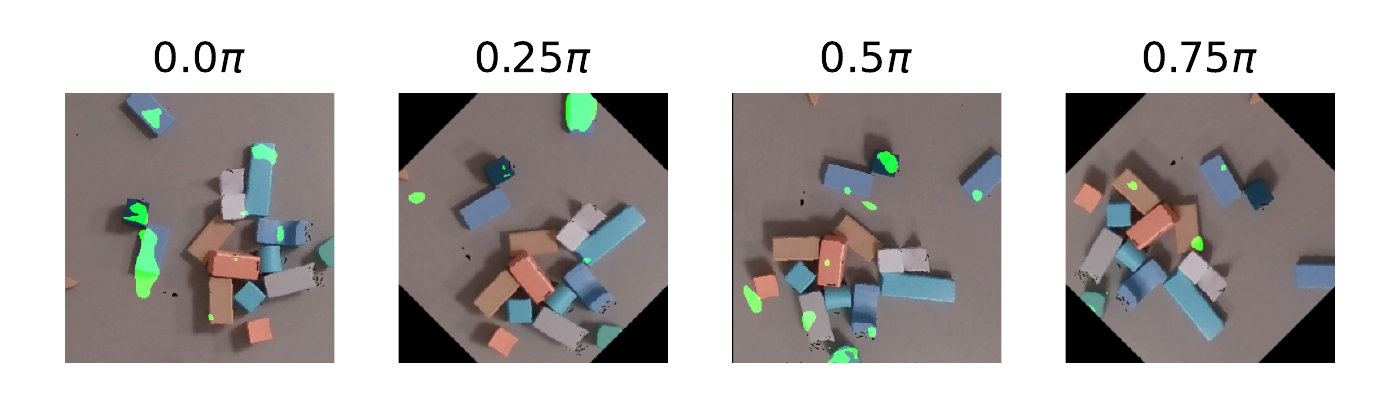}
    \end{center}
    \caption{Outputs of the affordance network, which will model the grasping environment. Highlighted regions are grasps that should result in grasp success, conditioned on a fixed horizontal gripper angle. The figures are rotations of the same scene. The model achieves 80\% balanced accuracy.} 
    \label{figure:affordances}
\end{figure}

We focus on vision-based grasping, where the input is a top-view image, and the goal is to generate pick-point predictions (see figures \ref{figure:affordances}, \ref{figure:teacher-student-outputs}). Many of the most effective grasping systems are trained on abundant datasets of real-world experience \cite{pinto2016supersizing}, \cite{levine2018learning}, \cite{kalashnikov2021mt}, \cite{kalashnikov2018qt}. Consequently, gathering real-world training data may quickly become a bottleneck. One potential route is to train purely in simulation~\cite{dexnet}, \cite{depierre2018jacquard}, however here we focus on a different scenario.

Imagine that we want to develop a robot picking system that will grasp items from bins in a warehouse. It is easy to install a camera that will take a photo of every bin passing underneath a conveyor belt, generating large quantities of images. Those images belong to our target domain - we want to generate pick-point predictions for the products present in the images. The only problem is that we do not have any signal (like robot experience) that we could use to train our models. The goal of this paper is the following: \textbf{use a limited amount of real robot experience} (grasping attempts) to train a model effectively by \textbf{taking advantage of an abundant dataset of unlabeled images}.

\subsection{Semi-supervised learning}
Semi-supervised learning is a type of machine learning where a model is trained on a combination of labeled and unlabeled data. Labeled data refers to data that has already been categorized or classified, while unlabeled data refers to data that has not yet been labeled.

Semi-supervised learning is particularly useful when acquiring labeled data is time-consuming or expensive, but plenty of unlabeled data is available. This is exactly the case
described in our warehouse robot example. By training on a combination of labeled and unlabeled data, a model can learn to recognize patterns in the data and make predictions with a higher degree of accuracy.

In semi-supervised learning, the labeled data is used to guide the model's learning process, while the unlabeled data helps the model to generalize and make predictions on new, unseen data.

\subsection{Contributions}
The main contribution of this work is a student-teacher-based semi-supervised learning algorithm suited for robotic picking, which can leverage large amounts of unlabeled data obtaining substantially better results compared to the purely fully-supervised learning setting. In the regime of a small number of robot training
samples, taking advantage of the unlabeled data allows us to
achieve performance at the level of 10-fold bigger dataset size
used by the baseline, see figure~\ref{figure:teacher-student}.

We introduce a new top-down grasping dataset and a new grasping simulation environment. 

We perform extensive tests validating the superiority of our method over the baseline. The most reliable metric is generated by using over 10 thousand grasp attempts on the real robot, but to ablate our results and understand the influence of various factors we also use proxy metrics: neural-network based grasp-success proxy, grasping simulation environment.

We release datasets, code for training, simulation environment and real environment at \url{https://github.com/nomagiclab/grasping-student}.

\section{RELATED WORK}
While we are unaware of attempts to use a semi-supervised approach in robotic grasping, using the additional dataset of unlabeled samples was beneficial in many other domains.

\subsection{Semi-supervised learning algorithms in Computer Vision}
Semi-supervised learning already took-of in the computer vision space, producing algorithms that were state-of-the-art in many computer vision tasks.
FixMatch \cite{sohn2020fixmatch} achieved state-of-the-art performance across a variety of image classification benchmarks.
Noisy student \cite{xie2020self} and Meta Pseudo Labels \cite{pham2021meta} are both semi-supervised approaches based on the student-teacher schema that have beaten the ImageNet challenge in the years of their release and
soft teacher \cite{xu2021end} achieved state-of-the-art on the COCO object detection task.

Generally speaking, semi-supervised algorithms for vision can be classified into the following two categories:

\subsubsection{Student–Teacher schema}
In this technique, a neural network model called the student
is provided with the signal based on the predictions of a base model or algorithm called the 
teacher. The aim of this technique is to expand the knowledge provided by the 
teacher through its predictions and activations on the unlabeled dataset:
\cite{sohn2020fixmatch}, \cite{xie2020self}, \cite{pham2021meta}, \cite{xu2021end}, \cite{grill2020bootstrap},
 \cite{zhang2021flexmatch}, \cite{tarvainen2017mean}, \cite{laine2016temporal}, \cite{caron2021emerging}.

\subsubsection{Self-supervised pretraining}
Another line of work that enables the leverage of unlabeled data is to learn the representations using an auxiliary objective that can be formulated without access to labels. Popular among those are contrastive pre-training methods, with its auxiliary objective being the discrimination between augmented versions of the same image and augmented versions of different images: \cite{chen2020simple}, \cite{zbontar2021barlow}, \cite{he2020momentum}. Another pre-training approach is BERT-like \cite{devlin2018bert} masked image modeling: \cite{zhou2021ibot}.

\subsection{Self-supervised methods in NLP}
Self-supervised pre-training allowed breakthroughs in the natural language processing domain.
It leverages the signal from the large unlabeled corpora of text and then uses the learned representations in downstream tasks. The main approaches are masked-language modelling \cite{devlin2018bert} and generative modelling \cite{radford2018improving}, \cite{radford2019language}, \cite{brown2020language}.

\subsection{Self-supervision in Reinforcement Learning}
Self-supervised methods are also successfully applied in Reinforcement Learning. In \cite{yu2022mask}, authors predict state representations from the observations with spatially and temporally masked pixels and show state-of-the-art performance on DeepMind Control Suite and discrete Atari benchmark.
Srinivas et al. in \cite{curl}  shows how to use contrastive learning on raw pixels to perform off-policy control on top of the extracted features extracts high-level features.
In \cite{seo2022reinforcement}, Seo et al. achieve state-of-the-art on many manipulation and locomotion tasks using generative pre-training on videos. 
Zhang et al. show how to leverage unlabeled trajectories to form useful visual representations, \cite{zhang2022light}.

\subsection{Self-supervised learning for robotics}
Self-supervision in the context of robotics often means the ability to collect data partially or completely autonomously. Grasp labels can be obtained by stochastic picking:
\cite{pinto2016supersizing}, \cite{levine2018learning}, \cite{kalashnikov2021mt}, \cite{kalashnikov2018qt}, \cite{grasp2vec}, \cite{zeng2018learning}, \cite{yen2020learning}. 
Navigation models can be trained by stochastic or human-guided robot exploration \cite{shah2021ving}, \cite{gandhi2017learning}.
The same goes with goal-conditioned robot policies \cite{andrychowicz2017hindsight}.

For us, however, this approach is not satisfactory, as generating (even autonomously) lots of data on a real robot quickly becomes a bottleneck. Consequently, we are focused on the case of using an additional set of unlabeled data that is generated without the use of a robot.

\section{PROPOSED APPROACH}
We propose the following algorithm to leverage the unlabeled data for robotic grasping:
We first train a teacher baseline network from a small amount of labeled grasp data through imitation learning similar to transporter nets \cite{zeng2021transporter}, but without the sequential nature and placing module.
We use imitation learning as a baseline because it can achieve good grasp quality using a small number of labeled examples.
Then we put large amounts of unlabeled scenes through this model to obtain pseudo-labels as an argmax grasp of the prediction. 
In addition to leveraging the unlabeled data, we can correct the labels in the existing dataset. 
Moreover, we can get better label quality or resolution – for example, we can discretize the angle space when feeding the scenes to the teacher model to a higher resolution than in the original labeled dataset.

\subsection{Problem representation}
In our setting, we restrict ourselves to top-down grasping.
We model the grasping problem as a task of finding the function $f_\theta:~\mathbb{R}^{h\times w}~\longrightarrow~\mathbb{R}^{h\times w \times a}$ from a scene state representation space to a parametrization of a probability distribution over $SE(2)$ grasp representation space. Afterwards, the optimal grasping point for a given scene $x \in \mathbb{R}^{h \times w}$ can be recovered as $\arg \max_{(r, c, \phi) \in SE(2)} f_\theta(x)(r, c, \phi)$.

\subsection{Imitation Learning dataset}\label{dataset}
To train the grasp proposal network, we use the dataset consisting of $(x, g, y)$ triples, where $x \in \mathbb{R}^{h \times w}$ is the scene representation as a top-down orthographically projected depth image and $g \in h \times w \times a$ is the grasp trial description and $y$ is an indication whether grasp $g$ in the scene $x$ succeeded.
Grasp description $g$ consists of a row and a col on the depth representation of the scene, along with an angle at which the robot would try to grasp the object. 
In this work, we use only a depth channel as an input, but the dataset contains both RGB and depth.

The dataset we have collected consists of two parts:
\begin{itemize}
    \item human-labeled picks (2 thousand samples),
    \item picks from robot experience (3 thousand).
\end{itemize}

The first part was collected using the pipeline described in section \ref{section:real}, but using human-in-the-loop to give grasp proposals that should result in a successful pick. Not all of the picks succeeded, but with the grasp description, we additionally have the $y$ label, so we can use unsuccessful attempts as negative examples when training networks in the imitation learning fashion.

The second part was collected using the same pipeline but with a grasp proposal network in the loop, using the model developed during the first research iterations over the teacher baseline.
We use all data in the dataset to train the grasp-success rate proxy, which will model the real environment, see the section \ref{subsection:affordance-network} for details.
We use the human-labeled part to train the teacher model and the second part as an unlabeled dataset for the grasping student algorithm, after dropping the grasp-point information.

The size difference between the labeled and unlabeled parts of our dataset is smaller than in the standard self-supervised setting. We will use small subsets of the labeled dataset in the ablations and experiments in the following sections.
Scaling up the unlabeled part to extreme levels is an exciting direction for further research.

\begin{figure}[thpb]
    \begin{center}
        \includegraphics[scale=0.11]{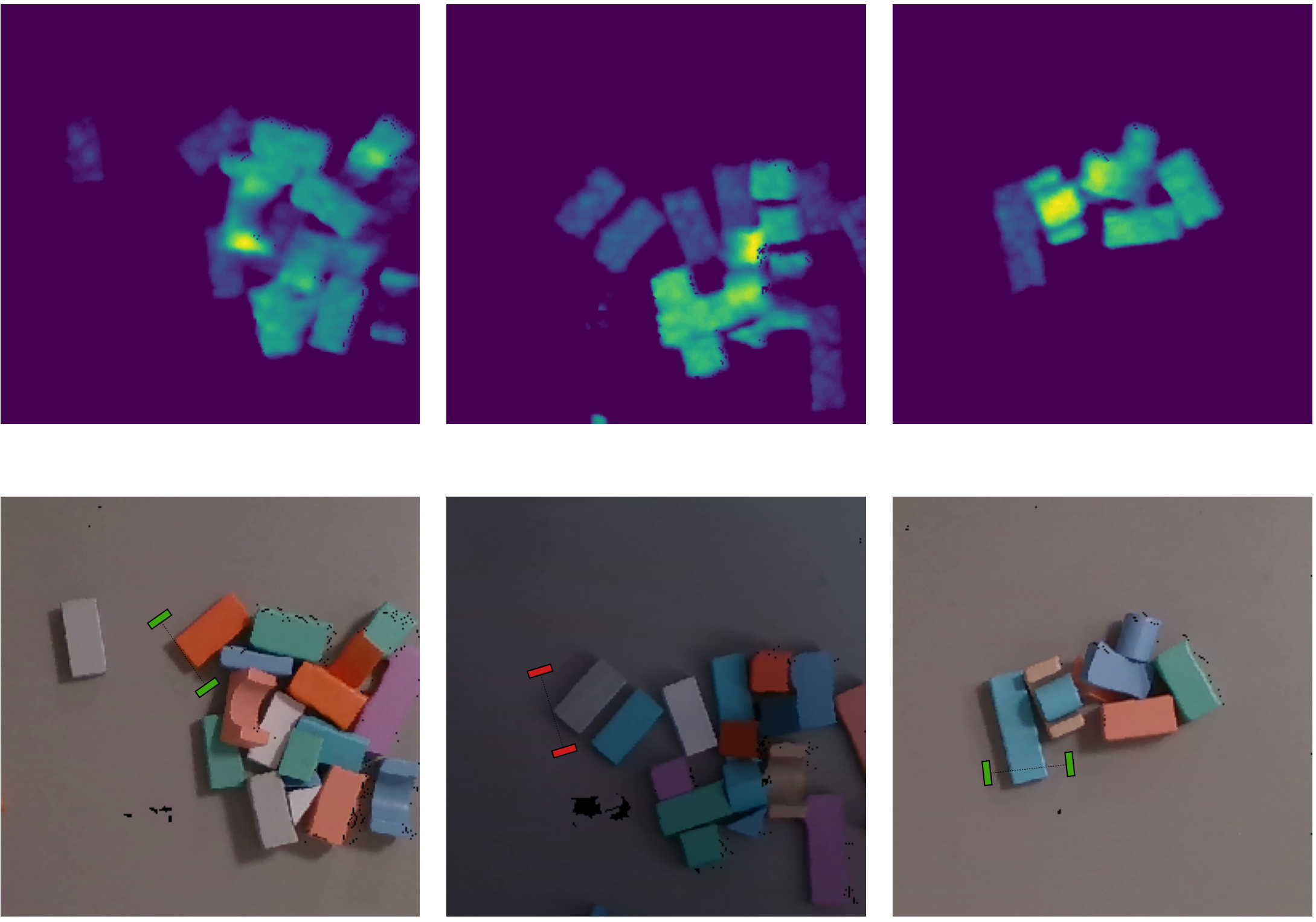}
    \end{center}
    \caption{Visualization of the three grasp examples from the imitation learning dataset. The top row shows depth visualizations, and the bottom shows corresponding RGB images. }
    \label{dataset-figure}
\end{figure}

\subsection{Training the teacher model}
To find the grasping proposal function: $f_\theta$, we use the behavioral cloning technique using the human-labeled part of the dataset \ref{dataset} to learn the auxiliary function $g_\theta:~\mathbb{R}^{h \times w}~\longrightarrow~\mathbb{R}^{h \times w}$, which will model our grasp proposal distribution $f_\theta(x)$ conditioned on a fixed angle. 

Later, $f_\theta$ can be reconstructed using $g_\theta$ as $$f_\theta(x)(r, c, \phi)~=~Rot_{–\phi}(g_\theta(Rot_\phi(x)))(r, c)$$ where $Rot_\phi$ is the plane rotation by angle $\phi$ around its center. The details are described in algorithm \ref{alg:teacher}.

\subsection{Angle space related optimization}\label{subsection:angle-optimization}
The standard technique is to divide the angle space $[0; \pi]$ into an equally spaced grid and put scene representation rotated by every angle on this grid through the network, see \cite{yen2020learning}, \cite{pickplace}, which is computationally expensive. We introduce optimization in training both the teacher and the student models. Instead of processing all of the angles from the grid through the network, we put only two angles through it: one from the labeled example and one distant from it.
We drop this optimization during the model deployment, where we densely sample the angle space.
The optimization refers to the $forward\_pair$ function used in the algorithms \ref{alg:teacher}, \ref{alg:student}.

\begin{algorithm}
\caption{Teacher learning through imitation}\label{alg:teacher}
\begin{algorithmic}[1]
\Require Labeled imitation dataset $D$
\Require Dataset $D$ to contain many successful grasps

\Procedure{forward}{$x, \phi$}
	\State \Return $Rot_{–\phi}(g_\theta(Rot_\phi(x)))$
\EndProcedure

\Procedure{forward\_pair}{$x, \phi$}
	\State $\phi' = \text{sample angle distant from } \phi$
	\State $p = $ \Call{forward}{$x, \phi$}
	\State $p' = $ \Call{forward}{$x, \phi'$}
	\State \Return $[p, p']$
\EndProcedure

\For{$(x, g, y) \in D$}
	\State $r, c, \phi = g$
	\State $[p, p'] = $ \Call{forward\_pair}{$x, \phi$}
	\State $[p, p'] = $ \Call{Softmax3D}{$[p, p']$}
	\State $l = $ \Call{BCE}{$p[r, c], y$}
	\State Update $\theta$ based on the gradient of $l$
\EndFor
\end{algorithmic}
\end{algorithm}

\subsection{Leveraging the teacher model to train the student model}
After training the base teacher in the imitation-learning setting on the human-picked part of the dataset \ref{dataset},
we can leverage it to use all the scenes without human-picked grasps.

We are processing all the scene representations through the teacher model to obtain these additional labels and taking the best-scored grasps as the pseudo-labels.
Finally, we train the student model on these pseudo-labels like we had trained the teacher.
The procedure is described in detail in the algorithm~\ref{alg:student}.

\begin{algorithm}
\caption{Student Learning}\label{alg:student}
\begin{algorithmic}[1]
\Require Labeled imitation dataset $D$
\Require Dataset $D$ to contain many successful grasps
\Require Large unlabeled dataset $U$
\State \text{Train the teacher model } $f'_\theta$ on $D$ using algorithm \ref{alg:teacher}
\For{$x \in U$}
	\State $r, c, \phi = \arg \max_{r, c, \phi \in SE(2)} g'_\theta(x)(r, c, \phi)$
	\State $[p, p'] = $ \Call{forward\_pair}{$x, \phi$}
	\State $[p, p'] = $ \Call{Softmax3D}{$[p, p']$}
	\State $l = $ \Call{BCE}{$p[r, c], 1$}
	\State update $\theta$ based on the gradient of $l$
\EndFor
\end{algorithmic}
\end{algorithm}

\subsection{Affordance Network as the grasp-success proxy}\label{subsection:affordance-network}
To be able to iterate over the algorithm and perform the ablation studies quickly, we are not evaluating each trained model directly in the environment. 
Instead, we train the affordance model, which models the probability of success of a grasp in a particular scene. 
We train it similarly to the teacher model. However, we use all available robotic experience. We do not use the $softmax$ at the end nor sample the additional angle contrastively as described in the subsection \ref{subsection:angle-optimization}.
Instead, we just put the rotated scene representation through the network and normalize the scores using the $sigmoid$ function on each pixel.
When training the affordance network, we resample the scenes and grasps to make the ratio of successful to unsuccessful grasp equal.
The above procedure is similar to works \cite{zeng2018learning}, \cite{yen2020learning}. Details in the definition of the algorithm \ref{alg:affordance-learning}.

\begin{algorithm}
\caption{Training the affordance network}\label{alg:affordance-learning}
\begin{algorithmic}
\Require Large labeled dataset $D$
\Require Dataset $D$ to contain many successful and failed grasps

\For{$(x, g, y) \in D$}
	\State $r, c, \phi = g$
	\State $p = $ \Call{forward}{$x, \phi$}
	\State $p = $ \Call{sigmoid}{p}
	\State $l = $ \Call{BCE}{$p[r, c], y$}
	\State Update $\theta$ based on the gradient of $l$
\EndFor

\end{algorithmic}
\end{algorithm}

\subsection{Training details}
We train our networks using the Adam optimizer \cite{adam} for 30 epochs. We use the learning rate of $10^{-3}$ with batch size 8 for teacher and student training and 16 for affordance network training — the largest that would fit on 8GiB memory GPU. We use $10^{-5}$ weight decay.
As the backbone networks, we use FPN \cite{fpn} resnet50 \cite{resnet50} for teacher and student models, U-Net \cite{unet} with resnet50 for affordance network that is used as a proxy for calculating the grasp success rate. 
As a learning rate schedule, we use the cosine schedule with minimal learning rate being $2 \cdot 10^{-6}$.
For model architectures and backbones implementations, we use PyTorch segmentation models library \cite{segmentation-models}, and for training them, we use PyTorch with PyTorch Lightning \cite{pytorch} \cite{pytorch-lightning}.

As the labeled imitation dataset $D$, we use human-selected picks from the dataset described in the section \ref{dataset}. As the unlabeled dataset $U$ for the student training, we use part of the dataset \ref{dataset} gained from robotic experience, size of $D$ is 2k, size of $U$ is 3k, 5k in total, but we vary the size of the $D$ as the ablation study.

We use all the available robotic experience from \ref{dataset} for training the affordance network.
We use only human-selected picks from \ref{dataset} to train the teacher model. We use all the available scenes in \ref{dataset} to train the student model, but without the corresponding labels.

We ensure fair evaluation and scores by doing proper train/test splits.
\section{EXPERIMENTS}

\subsection{Evaluation protocol}
We report grasp success rate as the proportion of the argmax-selected grasp-points from a given network that are classified by affordance network as a success – i.e., affordance score after sigmoid greater than 0.5. We report the exponential moving average of this metric on the validation dataset calculated over all 30 epochs, with $\alpha = 0.8$.

We condition the affordance model on the already processed angle to make the joint training and evaluation process quicker. It is provided from the dataset in the case of the teacher training and from the teacher in the case of the student training. Therefore, we only take the argmax on row and col, with a fixed angle, when calculating the grasp success rate proxy.

It approximates an actual grasp success rate score that would be achieved in a real-life system. Nevertheless, in further sections, we confirm our findings about the effectiveness of the grasping student method both in the simulation \ref{section:simulation} and the real-setup \ref{section:real}.

\subsection{Scaling the number of labeled examples}
We benchmark the grasping student method against the imitation learning baseline. 
We train both models similarly as described in algorithm \ref{alg:teacher}. 
In a nutshell, a grasping student is an imitation learning from labels obtained from a teacher model trained on (part of the) labeled data.

\begin{center}
\end{center}

\begin{figure}[thpb]
    \begin{center}
        \includegraphics[scale=0.5]{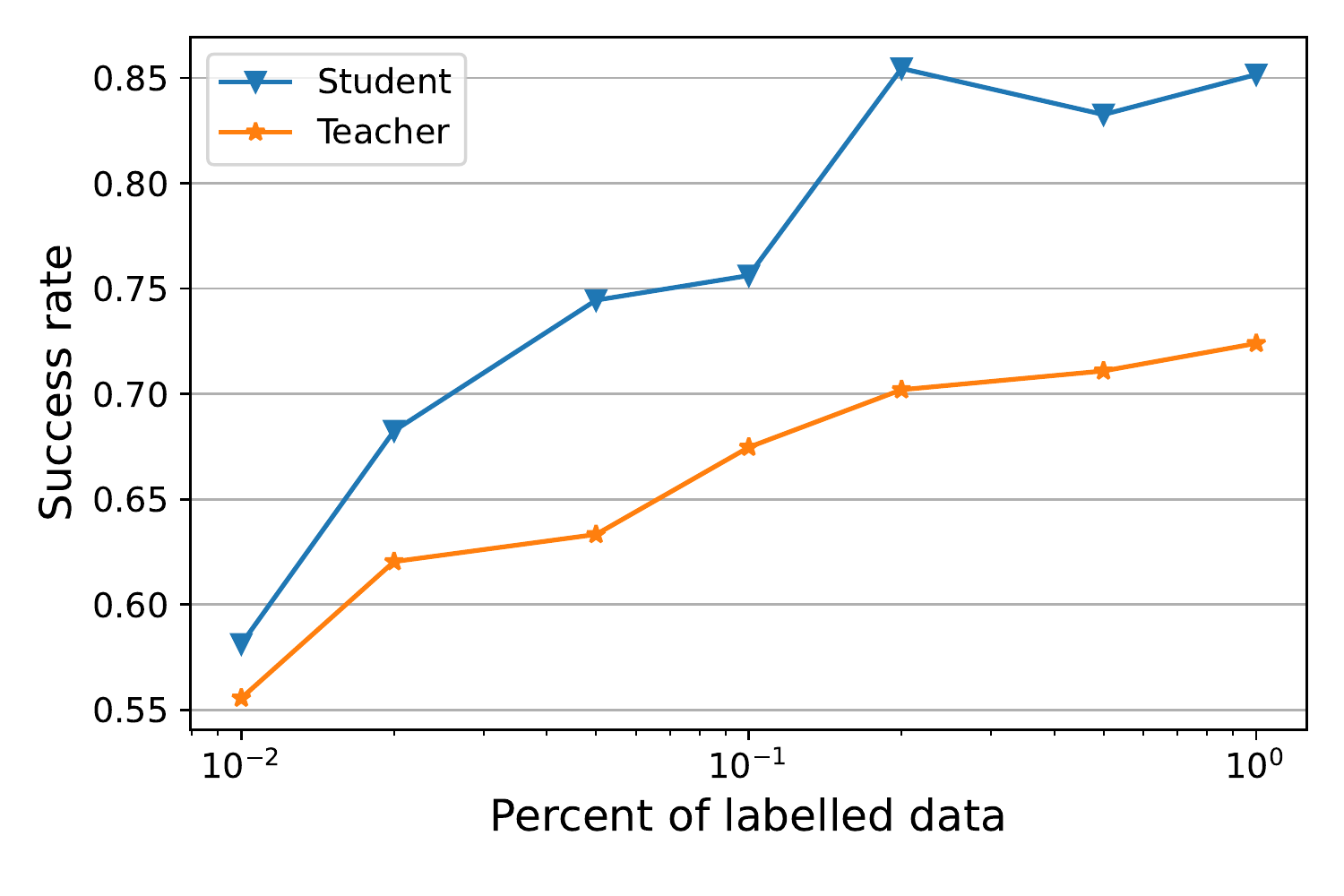}
    \end{center}
    \caption{Comparison between the grasp success rate of a teacher model and a student model based on the number of labeled examples used to train the teacher model. The student model outperforms the baseline by large margins. Grasping student with $5-10\%$ of the labels obtains greater performance than the baseline with all the available labeled data.}
    \label{figure:teacher-student}
\end{figure}

\subsection{Low entropy predictions of the student model}
\begin{figure}[thpb]
    \begin{center}
        \includegraphics[scale=0.6]{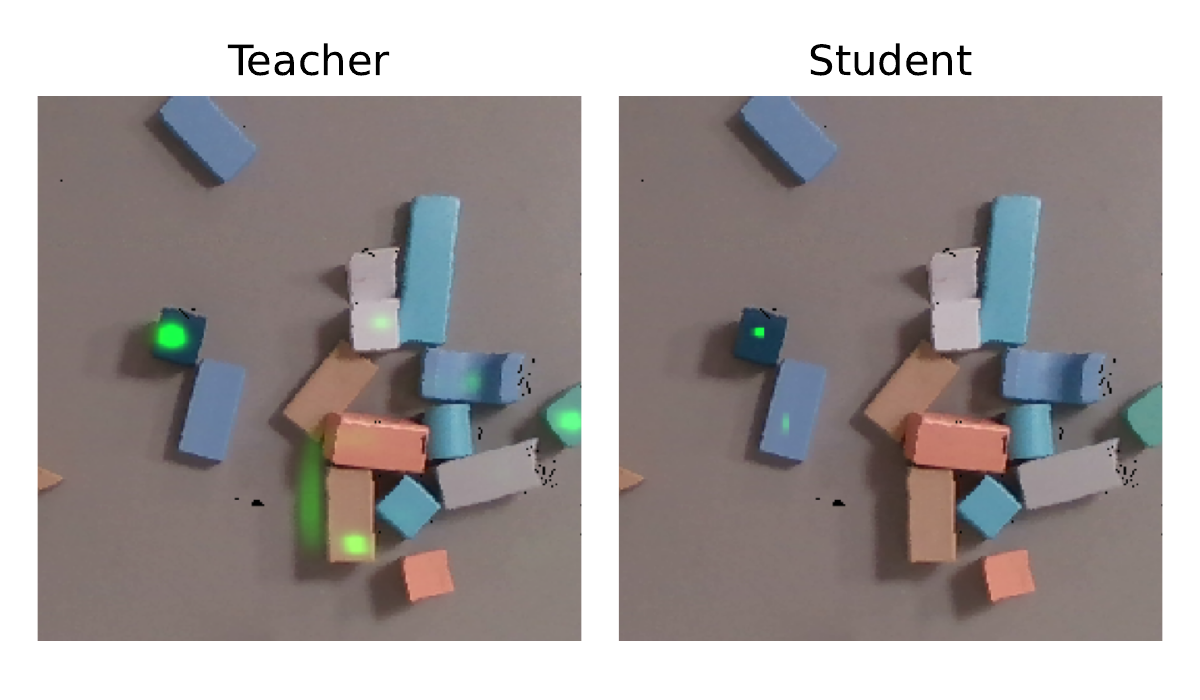}
    \end{center}
    \caption{Outputs of a teacher's and a student's models. The teacher model has a wider range of areas of interest than the student model. Both outputs are conditioned on a fixed horizontal gripper angle. Let us note that the student model outputs only highly robust grasps proposals compared to the baseline.}
    \label{figure:teacher-student-outputs}
\end{figure}

Due to how the student was trained, the entropy of its outputs is extremely low.
It is because taking labels as an argmax from the teacher model is much more predictable than a human grasp proposal, which has some inherent variability.
It is illustrated qualitatively in figure \ref{figure:teacher-student-outputs}.

\subsection{Angles when obtaining labels from teacher}

When obtaining the pseudo-labels from the teacher, we sample 16 angles uniformly from the equally spaced grid of $[0; \pi]$ of size 64. 
Increasing the number of samples when obtaining the labels from the teacher would result in labels of better quality, trading for processing time than was used to obtain a single label. 
Processing time naturally grows linearly. 
We performed the ablation experiment to confirm that an increased number of angles samples in student training results in a better-quality model.
We compare the grasp success rates proxies of models trained in the grasping student paradigm using various angles samples from the teacher network: 1, 2, 4, 8, 16, 32, 64. 
The grasp success rate grows with the number of angles sampled. The difference in the metric between the worst setting (1 angle sampled) and the best setting (all 64 angles on the grid chosen) is 30.32\%.

\begin{figure}[thpb]
\begin{center}
\includegraphics[scale=0.5]{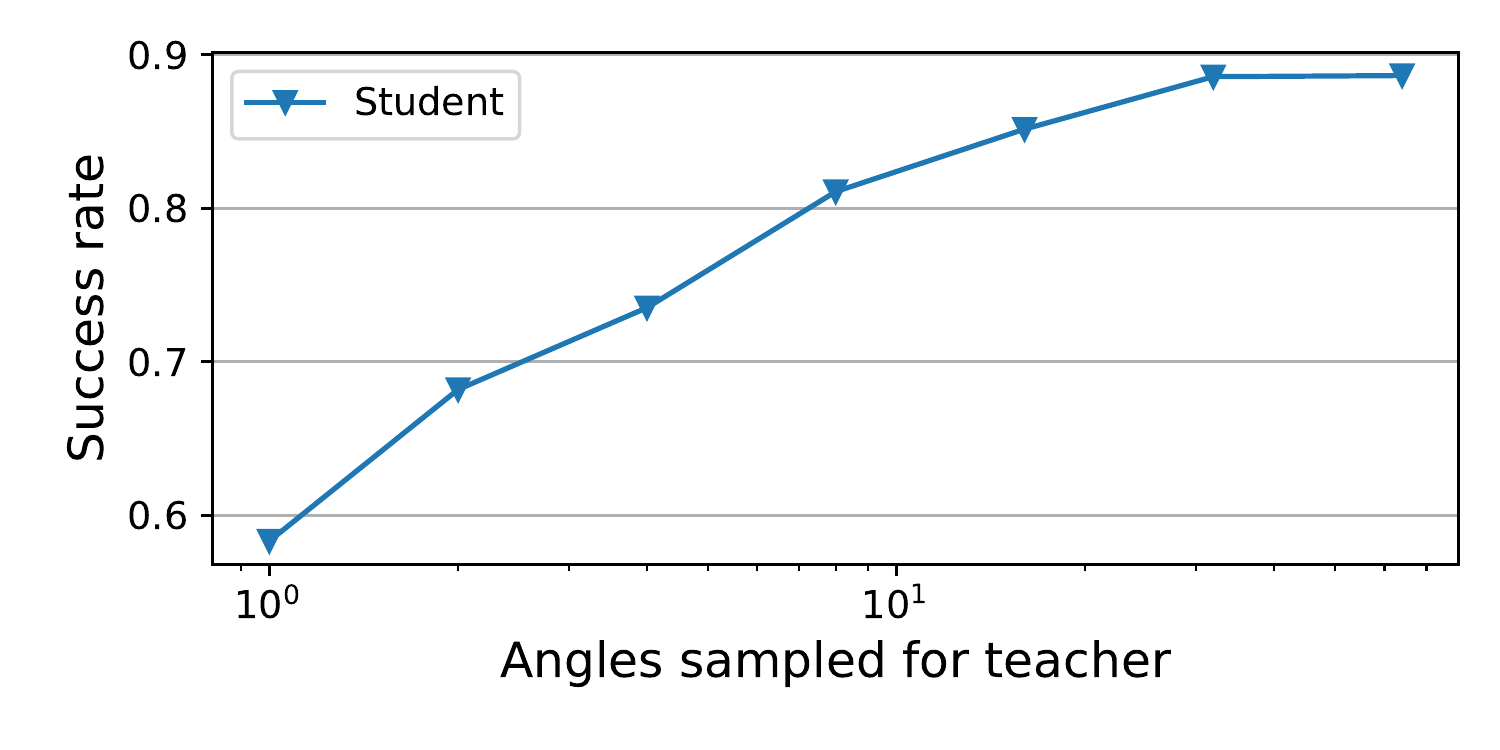}
\end{center}
\caption{Dependence of the success rate of the student model on the number of angles samples from the teacher. The success rate drastically increases with the number of angles sampled, reaching a 30.32\% difference.}
\label{figure:student-angles-sampled}
\end{figure}

\subsection{Top-n teacher labels in student training}
To obtain pseudo-labels from the teacher, we use the argmax of the prediction scores for each grasp. We can get multiple grasp proposals from each scene, not only the one with the highest teacher score.
We try the following idea: after obtaining the grasp with the highest teacher score, we ban grasps in proximity and select the next highest score grasp. We can repeat this procedure many times, treating each grasp selected as a new label for the student model. 
We investigate it by varying the $n = 1, 2, 3, 4$ and observe that the quality of the student model deteriorates drastically with $n$ increasing.
We believe that one should be able to modify the algorithm to obtain more than one grasp label from the scene with a positive impact on model quality – for example, by weighing the grasp proposals, taking into account the teacher score or other grasp features. We leave it as a further exciting research question.

\begin{figure}[thpb]
\begin{center}
\includegraphics[scale=0.5]{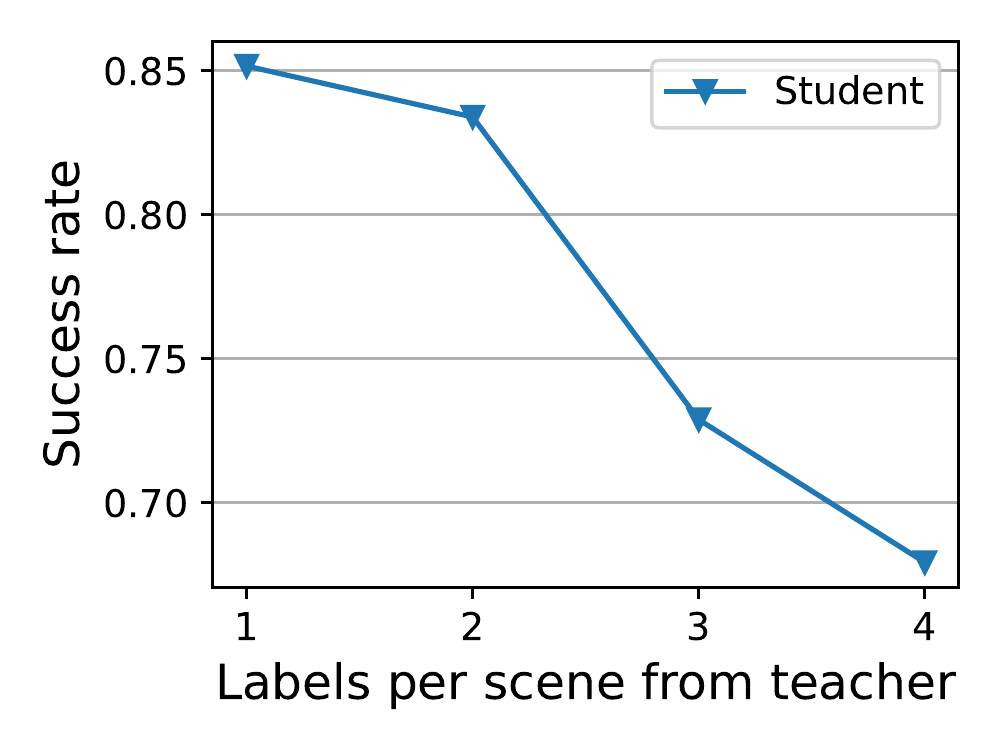}
\end{center}
\caption{Influence of the number of labels obtained from the teacher in a top-on fashion to the student's success rate for $n = 1, 2, 3, 4$. It quickly deteriorates with increasing $n$, reaching a 17.24\% success rate difference.}
\label{figure:student-topn}
\end{figure}

\section{SIMULATION EVALUATION}\label{section:simulation}
We confirm our findings about the effectiveness of the grasping student method in the simulation setup. 
We imitate the real setting as closely as possible, having simulated a two-finger gripper with a stroke close to the stroke in the real setting and blocks similar to the real setting. 
We use the same code for the evaluation in the simulation, and in the real world, we only swap the environment. Details on the evaluation procedure are described in section \ref{section:real}.
Note that this evaluation in the simulation is out of distribution testing because we trained the models using real-world data and experience.

\subsection{Setup}
We use seven different types of blocks of different shapes. Initially, we also additionally used the triangle blocks (8 types). However, in both the teacher and student networks, we could not learn how to pick it correctly (primarily due to inaccuracies of depth in the dataset \ref{dataset}).
Similarly, we have removed the triangular blocks from the real-life setup.

For simulator implementation, we use pybullet engine \cite{pybullet}.
We implement the two-finger gripper as five virtual prismatic joints and one revolute joint, three for moving freely in xyz, two for closing the fingers, and one revolute joint for gripper orientation. 

\begin{figure}[thpb]
\begin{center}
\includegraphics[scale=0.06]{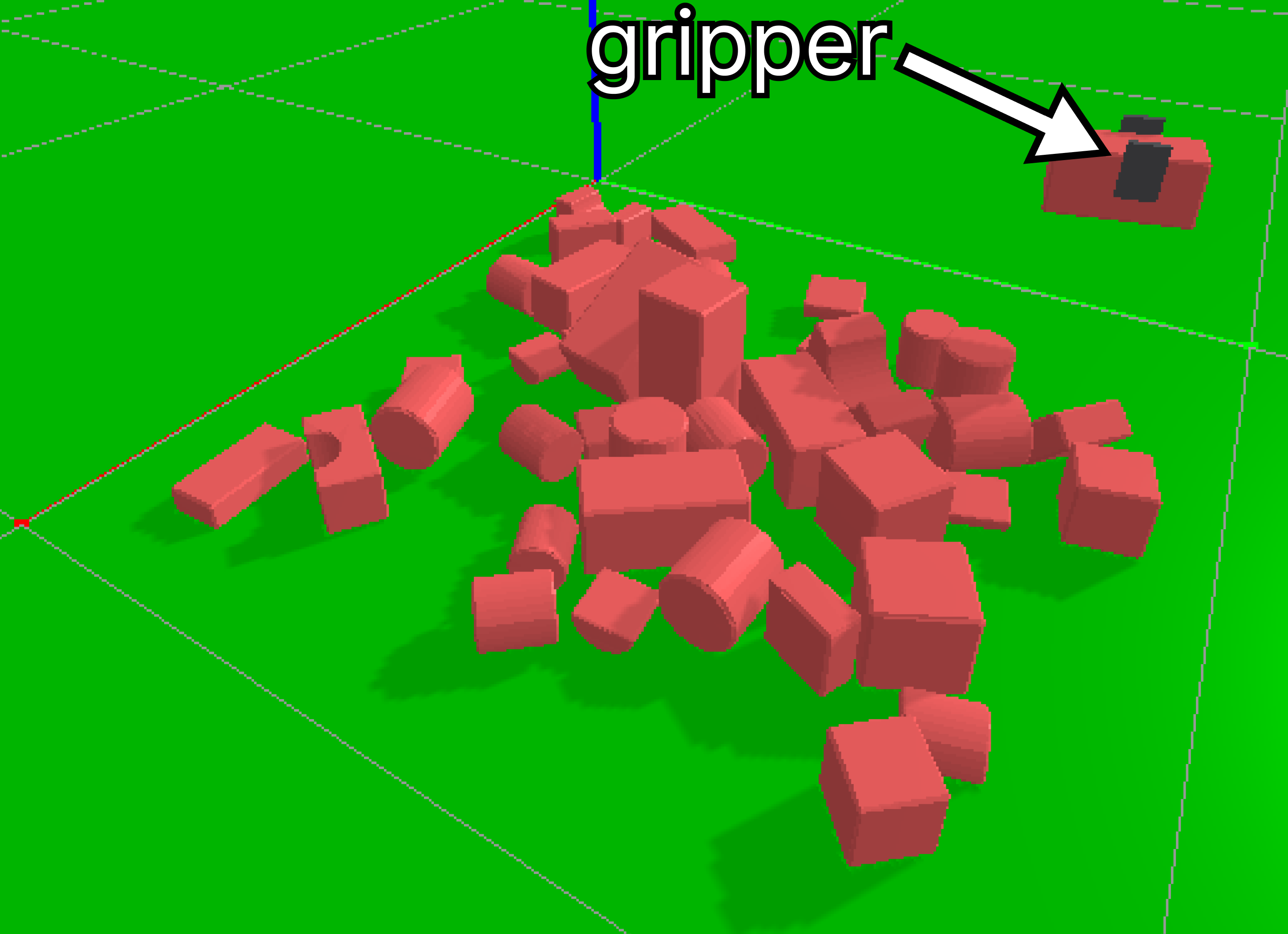}
\end{center}
\caption{Grasping in the simulation using the pybullet engine.
A simple two-finger gripper, implemented using a combination of prismatic and revolute joints, is trying to grasp seven types of blocks of diverse shapes, guided by a grasp proposal network and heightmap images calculated using the simulated camera.}\label{simulation-setup-figure}
\end{figure}

\subsection{Evaluation results}
We spawn 50 blocks, each of a random one of 7 types, with random scaling, orientation, and position in the scene. We try picking the blocks up continuously until the scene becomes empty or after five failures in a row. Then we reset the simulation, similarly as in section \ref{section:real}.
We repeat 25 thousand picks for each model: grasping student and the teacher.
We use the ones trained on 20\% of the labeled data because, in figure \ref{figure:teacher-student}, we have the largest gap between the teacher and a student model. We use the same models in the real-life setting \ref{section:real}.
We report a grasp success rate along with 95\% confidence interval for each model. 
\footnote{We use standard Wilson score interval method, assuming independence of grasp attempts and grasp success being a Bernoulli with fixed $p$.}
Results are presented in the table \ref{table:simulation-results}, with grasping student outperforming the baseline. 
Note that the absolute value of grasp success rate is insignificant because we could artificially make it higher for both methods by rendering easier scenes. The difference is the interesting bit. The same comment applies to the real \ref{section:real}.

\begin{table}[h]
\caption{Grasping in the simulation}
\label{table_example}
\begin{center}
\begin{tabular}{|c|c|c|}
\hline
& Imitation learning baseline & Grasping Student\\
\hline
Grasp success rate & $60.05\% \pm 0.61\%$ & $67.69\% \pm 0.58\%$ \\
\hline
\end{tabular}
\end{center}
\label{table:simulation-results}
\end{table}

\section{REAL-WORLD EVALUATION}\label{section:real}
In the real-world setup, we use the UR5e robot arm with a 2F-85 Robotiq gripper and the RealSense D415 camera fixed above the table. The setup is very similar to and heavily inspired by \cite{zeng2018learning}.
To automatically reset the scene and generate cluttered groups of items, we use two large boxes standing on the table next to each other in a fixed position and a single smaller box with movable bottom positioned inside the large right box. In all our experiments, we use random arrangements of toy blocks of different shapes and colors. Each scene consists of 1-39 such blocks, most within the area available for grasping.
\begin{figure}[thpb]
    \begin{center}
        \includegraphics[scale=0.27]{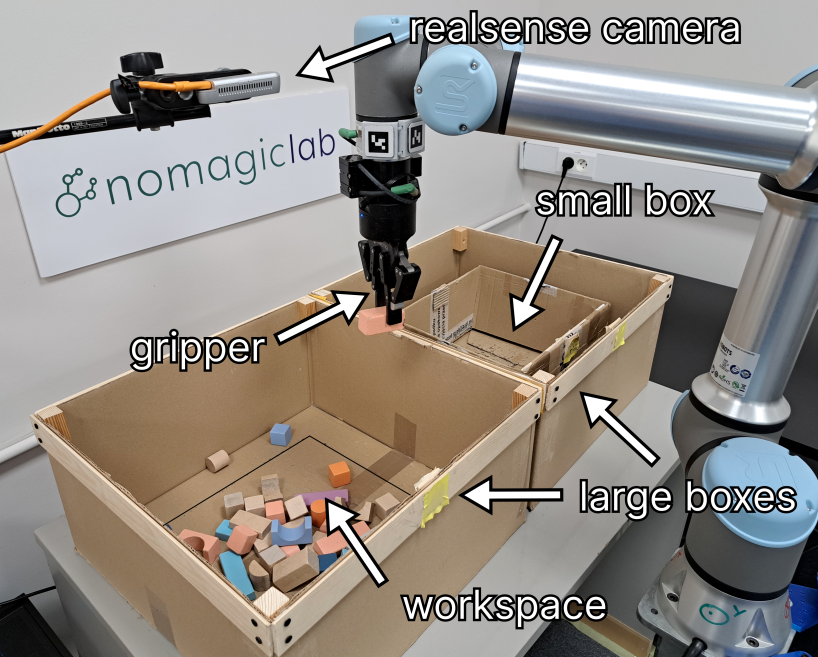}
    \end{center}
    \caption{The grasping environment consists of the camera, grasping robot, and three boxes. The workspace is an area inside the large left box used by the networks to predict the best grasping positions and angles.} 
    \label{figure:real}
\end{figure}

\subsection{Workspace}
The workspace is the square area of size 45cm x 45cm located inside the left box and represents the part of the scene visible to the networks. To avoid collision of the gripper with the walls of the large box, we add margins of 6 cm on each side, where the gripper ignores affordances from the networks and is not allowed to grasp.

\subsection{Epochs and resets}
The experiments are conducted in epochs, where every epoch starts with resetting the scene. Automatic reset is possible by manipulating the boxes with the gripper, where the small box with the movable bottom is used to ensure the scenes are cluttered. 

The grasping in a single epoch consists of moving the blocks from the workspace and dropping them to the large right box until the part of the workspace available for grasping is empty or there are five subsequent unsuccessful grasps. We assume the scene is empty if at least 98\% of depth pixels equals the background depth.
We evaluate the teacher on 204 epochs (5576 grasping attempts total) and the student on 202 epochs (6048 grasping attempts total). To ensure the environment consistency between student and teacher, we conduct the experiments alternatingly, where an epoch of testing the teacher always precedes every epoch of testing the student.

\subsection{Heightmaps}

We use the heightmaps generated from the RGB-D images captured by the camera to evaluate the networks. Heightmaps are orthographic projections of the images onto the table cropped to fit the workspace. They are generated from a 3D point cloud obtained from the image captured by the camera (1280 × 720 RGB-D images).
The heightmaps cover the 45cm x 45 cm rectangle, which translates to 338 x 338 pixels. The area available for grasping (after ignoring the margins) is a rectangle of 33cm x 33cm, equivalent to 248 x 248 pixels.
In the experiments, we use only depth and ignore RGB channels.

\subsection{Grasping policy and results}
We use neural networks with the argmax strategy to select the best grasping position and an angle ($r, c, \theta$) using the same architecture as in training. There are 16 possible angles. The height at which the gripper is lowered during the grasp attempt is calculated by taking the average of detected depth on a 3x3 pixel square around $(r, c)$. We set the stroke of the fingers to the maximum (85mm) before every grasp.

The results of the experiments in a real-life setting are summarized in table \ref{table:real}. There is a clear improvement in grasp success rate when using the student network, confirming our findings about the grasping student efficiency.
\begin{table}[h]
\caption{Grasping in the real world}
\label{table:real}
\begin{center}
\begin{tabular}{|c|c|c|}
\hline
& Teacher & Student\\
\hline
Grasp success rate & $65.25\% \pm 1.25\%$ & $78.64\% \pm 1.04\%$\\
\hline
\end{tabular}
\end{center}
\end{table}

\section{CONCLUSIONS}
By successfully applying a semi-supervised learning approach to robotic grasping, we have achieved better results than the standard fully-supervised baseline. Consequently, one can obtain better performance with the same robot experience or the same performance using less robot experience. 

Our algorithm is based on an already established teacher-student scheme widely used in semi-supervised computer vision tasks. We have shown that the semi-supervised methods may also be applied in a robotics domain, which should greatly benefit from this type of solution because of the hardship of obtaining real robotics experience.

Moreover, we have additional findings about our grasping student algorithm: the quality improves with the number of angles we feed through the teacher model,
the metrics deteriorate with increasing the number of labels we obtain from a teacher from a single image in a top-n fashion,
the student model is much more confident in their predictions than the baseline model.

An exciting avenue for future research is applying other semi-supervised learning schemes, such as contrastive pretraining, self-distillation, and others. It is known that the most effective augmentation used in self-supervised contrastive pretraining is cropping \cite{chen2020simple}, which is not directly applicable in the robotic picking setting, consequently making the applicability of contrastive pretraining non-trivial.

An interesting task is obtaining a good-quality grasping model using a small number of labeled examples (about 100 or less) using large volumes of unlabeled scenes.
Another research direction may be experimenting with domain adaptation through the teacher-student procedure.
Making robot learning more data-efficient is an exciting and essential line of work that requires further research.

\section*{APPENDIX}
In tables \ref{table:teacher-student}, \ref{table:student-angles-sampled}, \ref{table:student-topn}, we show the precise quantitative results of the experiments, scored by a proxy affordance network.
They refer to the qualitative results from \ref{figure:teacher-student}, \ref{figure:student-angles-sampled}, \ref{figure:student-topn}.

\begin{table}[h]
\caption{Varying the amount of labeled data, see fig. \ref{figure:teacher-student}}
\label{table:teacher-student}
\begin{center}
\begin{tabular}{|c|c|c|c|}
\hline
Percent of labeled data & Teacher & Student & Difference\\
\hline
0.01 & 55.56\% & 58.13 \% & 2.57\%\\
\hline
0.02 & 62.04\% & 68.25\% & 6.21\%\\
\hline
0.05 & 63.33\% & 74.45\% & 11.12\%\\
\hline
0.1 & 67.47\% & 75.62\% & 8.15\%\\
\hline
0.2 & 70.20\% & 85.45\% & 15.25\%\\
\hline
0.5 & 71.09\% & 83.26\% & 12.17\%\\
\hline
1.0 & 72.40\% & 85.16\% & 12.76\%\\
\hline
\end{tabular}
\end{center}
\end{table}

\begin{table}[h]
\caption{Varying the number of angles sampled, see fig. \ref{figure:student-angles-sampled}}
\label{table:student-angles-sampled}
\begin{center}
\begin{tabular}{|c|c|c|c|}
\hline
Angles sampled & Student success rate \\
\hline
1 & 58.33\% \\
\hline
2 & 68.20\%\\
\hline
4 & 73.53\% \%\\
\hline
8 & 81.09\% \\
\hline
16 & 85.16\% \\
\hline
32 & 88.58\% \\
\hline
64 & 88.65\% \\
\hline
\end{tabular}
\end{center}
\end{table}

\begin{table}[H]
\caption{Varying the number of labels from a single scene, see fig. \ref{figure:student-topn}}
\label{table:student-topn}
\begin{center}
\begin{tabular}{|c|c|c|c|}
\hline
Labels per scene & Student success rate \\
\hline
1 & 85.16\% \\
\hline
2 & 83.39\%\\
\hline
3 & 72.88\% \%\\
\hline
4 & 67.92\% \\
\hline
\end{tabular}
\end{center}
\end{table}

\subsection{Statement of work}
Piotr Krzywicki is the leading author responsible for the project, including the final formulation of the grasping student algorithm, performing experiments, and most of the implementation. Krzysztof Ciebiera performed experiments on the initial version of the semi-supervised approaches, contributed to the algorithm design, and set up the pipeline in simulation. Rafał Michaluk prepared the real-world setup (both hardware and software) that allowed for autonomous testing on the robot. Inga Maziarz prepared the dataset for imitation learning. Marek Cygan suggested using semi-supervised learning for grasping (without specific algorithm formulation) and supervised the project.

\section*{ACKNOWLEDGMENT}
The authors are thankful to Jan Dziedzic and Karol Pieniacy for their help with the hardware. Heavy computation was performed using the Entropy cluster funded by NVIDIA, Intel, the Polish National Science Center grant UMO2017/26/E/ST6/00622, and ERC Starting Grant TOTAL. Real-world experiments were performed in Nomagiclab at the University of Warsaw.

\bibliography{root} 
\bibliographystyle{IEEEtran}



\end{document}